\pdfoutput=1
\documentclass[11pt]{article}

\usepackage[]{acl}

\usepackage{times}
\usepackage{latexsym}

\usepackage[T1]{fontenc}
\usepackage[utf8]{inputenc}
\usepackage{microtype}
\usepackage{inconsolata}
\usepackage{amssymb}
\usepackage{amsmath}
\usepackage{amsfonts}
\usepackage{booktabs}
\usepackage{graphicx}
\usepackage{multirow}
\usepackage{makecell}
\usepackage{enumitem}
\usepackage[normalem]{ulem}
\useunder{\uline}{\ul}{}

\usepackage{inconsolata}
\newcommand{\inlineSubsection}[1]{
  \par\noindent\textbf{#1}\quad
}

\title{UniArk: Improving Generalisation and Consistency for \\ Factual Knowledge Extraction through Debiasing}

\author{Yijun Yang$^1$, Jie He$^1$, Pinzhen Chen$^1$, V\'ictor Guti\'errez-Basulto$^2$,\and Jeff Z. Pan$^1$  \\
         $^1$School of Informatics, University of Edinburgh, UK  \\  $^2$School of Computer Science and Informatics, Cardiff University, UK
           \\ \tt {thomasyyj@outlook.com, j.he@ed.ac.uk, pinzhen.chen@ed.ac.uk}\\ \quad {\tt gutierrezbasultov@cardiff.ac.uk,  j.z.pan@ed.ac.uk}
}

\begin{document}
\maketitle
\begin{abstract}
Several recent papers have investigated the potential of language models as knowledge bases as well as the existence of severe biases when extracting factual knowledge. In this work, we focus on the factual probing performance over unseen prompts from tuning, and using a probabilistic view we show the inherent misalignment between pre-training and downstream tuning objectives in language models for probing knowledge. We hypothesize that simultaneously debiasing these objectives can be the key to generalisation over unseen prompts.  We propose an adapter-based framework, \textbf{UniArk}, for generalised and consistent factual knowledge extraction through simple methods without introducing extra parameters. Extensive experiments show that UniArk can significantly improve the model's out-of-domain generalisation as well as consistency under various prompts. Additionally, we construct \textbf{ParaTrex}, a large-scale and diverse dataset for measuring the inconsistency and out-of-domain generation of models. Further, ParaTrex offers a reference method for constructing paraphrased datasets using large language models.\footnote{The ParaTrex dataset and code are available at \url{https://github.com/Thomasyyj/UniArk}.} 

\end{abstract}

\section{Introduction}
Pre-trained Language Models (LMs) have been widely adopted in the NLP field. A key reason for the uptake of LMs is their capability to store knowledge in the parameters learned through pre-training \cite{PRKS+2023,liu2023pre}. Many works have looked at how to treat LMs as knowledge bases by extracting and measuring knowledge graph triples  from them~\citep{PVGW2017,Pan2017b}. LAMA \cite{petroni-etal-2019-language} is the first benchmark for measuring the extracted factual knowledge from LMs. In LAMA, factual knowledge is represented as triples (\textit{subject, relation, object}) and is extracted through manually designed prompt templates. For example, to answer the query (\textit{Barack Obama, place of birth, ?}), we query LMs using the prompt: ``\textit{Barack Obama was born in [MASK]}''. 
\begin{figure*}[t]
\centering
\includegraphics[width=0.7\linewidth,height=0.4\linewidth]{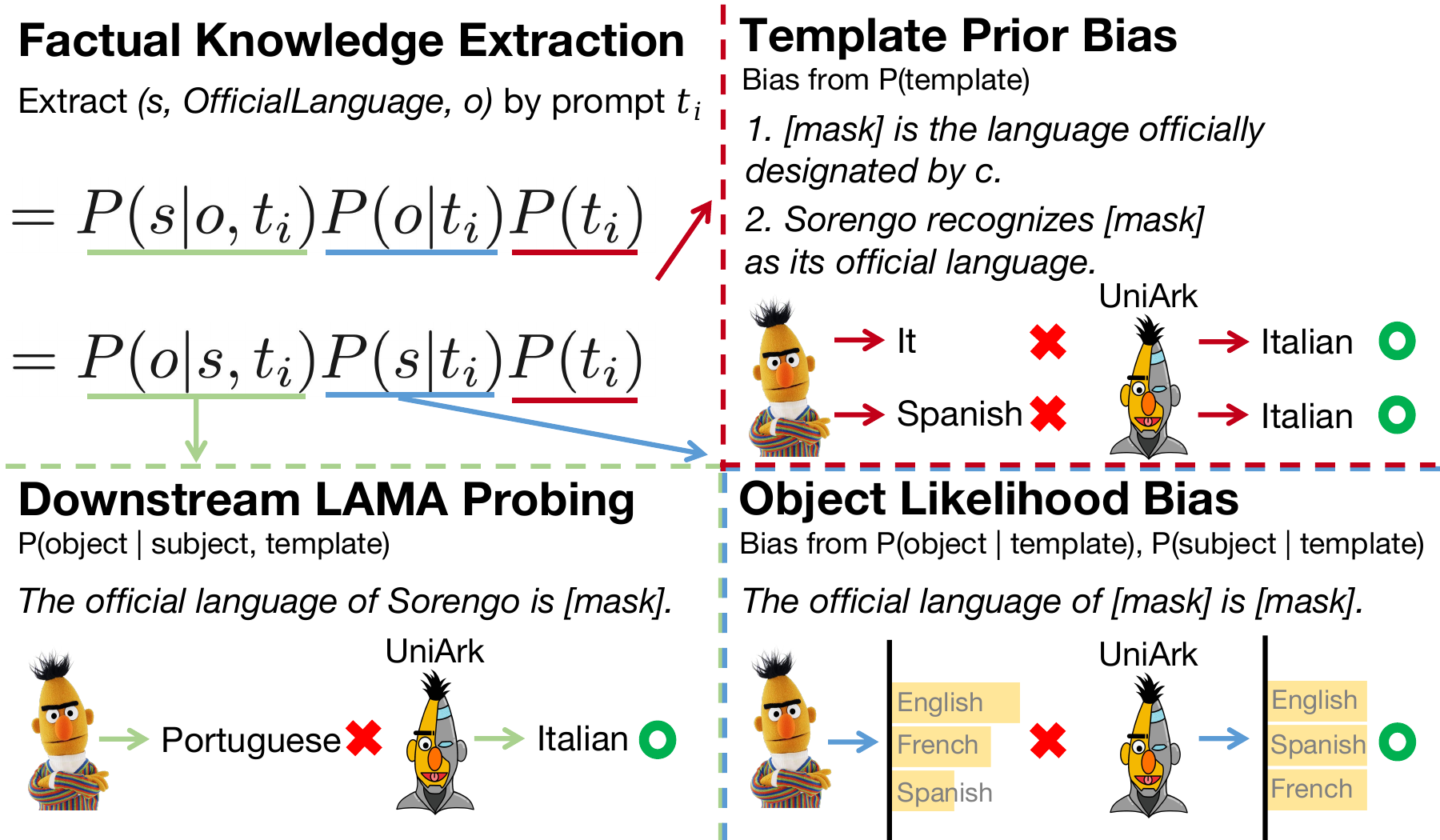}
\caption{Illustration of the inherent objectives' bias from the template prior and template verbalization, with a comparison to our UniArk framework.}
\label{intro}
\end{figure*}

Many subsequent works have searched for optimal prompting strategies in order to improve the accuracy of extraction ~\cite{shin-etal-2020-autoprompt,li-liang-2021-prefix,LIU2023,Li2022TaskspecificPA}. However, due to the limitation of LAMA, which only provides one prompt template for each relation,  they only tested prompts that LMs have seen during training, yet considered their consistency over different paraphrased prompt templates. On the contrary, \citet{elazar2021measuring} and \citet{newman2021p1} focused on the consistency between predictions from semantically similar prompts, but did not look at optimizing the models'  accuracy. In light of this, in this work we investigate how to improve both accuracy and consistency for  unseen prompt templates, i.e.\ out-of-domain generalisation. We perform a probabilistic decomposition of the factual knowledge retrieval objective $P(\textit{subject}, \textit{object}|\textit{relation})$, cf. Fig\ \ref{intro}, and find a misalignment between the pre-training and tuning objectives. This exposes two biases: $P(\textit{subject}|\textit{template}), P(\textit{object}|\textit{template})$ (bias from object likelihood) and $P(\textit{template})$ (bias from template prior) as shown in Fig \ref{intro}. Object likelihood bias refers to the likelihood of a predicted object given template-only prompts, such as ``\textit{The official language of [MASK] is [MASK]}'', being biased. The biased object likelihood has been shown to positively correlate with the predictions from subject-given prompts and negatively influence the performance of factual extraction \cite{wang-etal-2023-towards-alleviating, cao-etal-2021-knowledgeable}. Template prior bias is defined as the inconsistency among outputs from prompt paraphrases due to the domination of specific verbalizations during pre-training.

We propose \textbf{UniArk}, a parameter-free unifying framework  for optimizing both accuracy and consistency, through debiasing. The key idea behind each debiasing module is to equalize the probability distribution for the decomposed source bias term. To this end, we choose adapter-tuning as our base tuning method, which is widely accepted as a modular parameter-efficient way of tuning and an effective way of  debiasing~\cite{kumar-etal-2023-parameter,lauscher-etal-2021-sustainable-modular}. However, to the best of our knowledge,  we are the first to investigate adapter-tuning in factual knowledge probing tasks. 

To evaluate the performance under unseen prompt templates, a paraphrased benchmark of the LAMA dataset is needed. We argue that the existing dataset ParaRel \cite{elazar2021measuring} is both small in scale and not lexically diverse enough, as it is constructed based on rule-based methods such as swapping specific phrases. Therefore, we propose the dataset \textbf{ParaTrex} which is constructed using 
the large language model GPT-3.5. ParaTrex provides a more complex and substantially larger paraphrasing dataset. We provide both automatic evaluation and human evaluation statistics to show its high quality. Our main contributions are:

\begin{itemize}[noitemsep,topsep=-\parskip]

\item We focus on the out-of-domain generalisation of factual probing and point out the misalignment between the pre-training and tuning objectives in a probabilistic view, exposing the bias under a unified view as well as showing the possibility of improvements via debiasing.
\item We construct ParaTrex, a comprehensive benchmark   for out-of-domain generalisation measurements. We provide a thorough evaluation of ParaTrex.
\item We propose a simple and parameter-free method based on an adapter-tuning  framework for knowledge probing tasks.
Extensive experiments show the effectiveness of our methods in improving the generalisation performance of knowledge probing and mitigating biases. 

\end{itemize}

\section{Objective Decomposition}
\label{prob:decompose}
We  start with the objective for factual probing, showing that it is equivalent to the mask language modeling goals. We then decompose the probability representation of the task to show its misalignment with the tuning objectives, thus targeting two key components of the biased terms: the object likelihood and the template prior. We introduce several metrics for measuring these biased objectives.

Let $\mathcal{R} = \{r_1, r_2, \ldots, r_{n_r}\}$, $\mathcal{S} = \{s_1, s_2, \ldots, s_{n}\}$, and $\mathcal{O} = \{o_1, o_2, \ldots, o_{n}\}$ respectively  be sets of relations, subjects, and objects. 
Given a relation $r_j$, \emph{factual knowledge extraction} aims to extract factual knowledge triples ($s_i, r_j, o_k$) within LMs $\mathcal{M}$. Mathematically, we model $P(s_i, o_k|r_j)$ (the probability of subject-object pairs for a specific given relation). In practice,  we query $\mathcal{M}$ with a manually designed prompt template $t$ from the relation $r_j$. For instance, the template ``\textit{The capital of [X] is [Y]}'' is constructed from the relation ``\textit{Capital}''. Note that a specific relation can be mapped to different semantically similar prompt templates $\mathcal T = \{t_1, t_2, \ldots, t_{n_t}\}$. We predict $o_k$ through maximizing $P_\mathcal{M}(o_k|s_i, t_m)$.

To position the inherent misalignment when modeling the object probability, we use the following probability decomposition of the task objective:

{\small
\begin{align}
    &P(s,o | r)\\
= &\sum_{t_i \in \mathcal{T}}P(s,o,t_i)\label{eq:2} \\
= &\sum_{t_i \in \mathcal{T}}P(s,o | t_i)P(t_i)\label{eq:3}\\ 
= &\sum_{t_i \in \mathcal{T}}P(s | o, t_i)P(o | t_i)P(t_i)\label{eq:4}\\
= &\sum_{t_i \in \mathcal{T}}P(o | s, t_i)P(s | t_i)P(t_i)\label{eq:5}
\end{align}
}
Since  $\mathcal{T}$ is defined as the set of templates relevant to the relation $r$, we can drop $r$ in Eq.\ \eqref{eq:2}. We observe that the factual knowledge extraction goal $P(s,o|r)$ is equivalent to Eq.\ \eqref{eq:2}, which is approximated by the masked language modeling objective of LMs. After being decomposed, this objective function is influenced by five terms: $P(s | o, t_i)$, $P(o | s, t_i)$, $P(o | t_i)$, $P(s | t_i)$, and $P(t_i)$ (Eq.\ \eqref{eq:4} and Eq.\ \eqref{eq:5}). We note that sometimes we can rewrite object by subject since we might be interested in extracting the reversal relations, e.g. (\textit{United Kingdom, capital, London}) and (\textit{London, capital of, United Kingdom)}. The subject and object might therefore be substitutable for different relations on the same text corpus. We therefore treat $P(s | o, t_i)$, $P(o | s, t_i)$, and $P(o | t_i)$, $P(s | t_i)$ as the same in the remaining context. The first two terms coincide with our tuning objectives but additional terms are exposed, indicating that the objectives between pre-training and downstream tuning are not aligned. We refer to these additional terms as \emph{biased objectives}. $P(o | t_i)$, $P(s | t_i)$ show the bias from the object likelihood given a specific prompt template, $P(t_i)$ points out the bias from the template prior. 

\subsection{Bias from the Object Likelihood}\label{sec:promptprefbias}
We define the \emph{object likelihood} as $P(o|t)$. For  $t_k \in \mathcal{T}$, we then define the bias from the object likelihood as $P(o_i|t_k)\neq P(o_j|t_k)$ for all $ o_i, o_j \in \mathcal{O}$. That means that given only the prompt template without the subject, the object predicted by an LM is biased. This is also in line with the object bias defined in prior work \citep{wang-etal-2023-towards-alleviating}. To measure this bias, we propose the \emph{counterfactual hitting rate} (CT\_hit1). This measures the accuracy of outputs from the prompt-only inputs, which should be close to 0 due to the lack of subjects. We measure the bias from object likelihood on 4 types of popular tuning methods. 
Table \ref{table1} shows the average CT\_hit1 over all 41 relations in the LAMA dataset, where LAMA refers to do inference with the provided prompt in LAMA without tuning. Here we observe a clear increase in the hitting rate and entropy by comparing LAMA with other tuning methods, suggesting that after tuning, the model becomes stronger at guessing the correct answer from the likelihood of the object over the templates. 

To show the influence of the object likelihood bias over the accuracy of the prediction, we also report the Pearson correlation coefficient (R)  between the rank of grounding truth label over subject-given and subject-masked prompts over all samples in LAMA. In Table \ref{table1}, we can observe a positive correlation between object likelihood and subject-given predictions. Moreover, greater positive correlations are observed for the wrong cases. This implies that some of the inaccurate predictions can be attributed to the bias from the object likelihood.

\begin{table}[ht]
\centering\small
\begin{tabular}{lccc}
\toprule
                 & CT\_hit1 & R & R ($\times$) \\ 
\midrule
LAMA             & \phantom{0}5.23 & 0.322 & 0.353            \\
P-tuning         & 15.91 & 0.709 & 0.753            \\
Adapter          & 12.77 & 0.341 & 0.376            \\
Fine-tuning      & 13.11 & 0.228 & 0.284             \\ 
\bottomrule
\end{tabular}
\setlength{\belowcaptionskip}{-0.3cm}
\caption{Counterfactual hitting rates for prompt-only inputs and correlations (R) between the rank from outputs with and without given subject among all predictions and incorrect predictions (R ($\times$)).}
\label{table1}
\end{table}

\subsection{Bias from Template Prior}
The bias from the \emph{template prior} is defined as the inconsistency among different verbalizations with semantically similar prompt templates. Inconsistency problems have been widely discussed in previous works, \citep[][inter alia]{elazar2021measuring,newman2021p1}. This bias towards seen prompt templates $P(t_i)$ comes from unbalanced appearances of different prompts $t_i$ during pre-training. This will influence the quality of factual probing since the appearance of a specific prompt $t_i$ will weigh up $P(t_i)$, which results in learning better to predict $P(s,o|t_i)$ under this verbalization and neglecting other ones when being optimized. More importantly, this bias may be neglected in datasets such as LAMA where only one prompt template is used for tuning and testing. This motivates us to construct a more diverse and complex dataset for measuring the inconsistency as well as to propose a self-augmentation strategy aimed at averaging the biased template prior.

\section{The ParaTrex Resource}
We introduce the  \textbf{ParaTrex} resource, which is a large-scale and comprehensive paraphrasing dataset used for measuring both inconsistency and the generalisation capability of models on different unseen inputs. ParaTrex comprises 1526 paraphrases from 40 relations,\footnote{Like ParaRel \cite{elazar2021measuring} we omit one relation hard for generating paraphrases: ``[X] is a [Y]''}, with an average of 38.15 templates per relation. The statistics of the dataset are provided in Table \ref{table2}, with comparison to the ParaRel dataset  \cite{elazar2021measuring}. 

\label{sec3.3.1}
\begin{table}[tb]
\centering\small
\begin{tabular}{lcc}
\toprule
  & ParaRel & ParaTrex \\
\midrule
\# Relations & 39 & 40 \\
\# Patterns & 329 & 1526 \\
\midrule
Min \# patterns per rel. & 1 & 26\\
Max \# patterns per rel. & 20 & 47\\
Avg \# patterns per rel. & 8.23 & 38.15\\
Avg lexical per rel & 5.73 & 8.46\\
\bottomrule
\end{tabular}
\setlength{\belowcaptionskip}{-0.2cm}
\caption{Statistics of the ParaRel and ParaTrex datasets.}
\label{table2}
\end{table}

\smallskip
\subsection{Data Construction}
We construct ParaTrex, a paraphrased version of the LAMA dataset, using the following steps: (1) We begin with the patterns provided by LAMA. Each relation has one prompt template called \emph{base-pattern}. For example, the base pattern of relation "\textit{capital of}" is \textit{"[X] is the capital of [Y]."} (2) For each relation, to make the generation more specific, we extract its base pattern and its provided description corresponding to Wikidata \cite{wikidata}. 
For instance, for the relation \textit{CapitalOf}, "\textit{country, state, department, canton or other administrative division of which the municipality is the governmental seat}". (3) We formulate a manually crafted prompt directing GPT-3.5-turbo API to produce a total of 40 paraphrases. This includes 5 succinct paraphrases, each comprising no more than 7 words, as well as 5 extended paraphrases, each encompassing more than 15 words. More details of the paraphrase generation process can be found in Appendix \ref{Paratrex:construct}. (4) Through human inspection, we remove inappropriate paraphrases characterized by excessive ambiguity or similarity to preceding generations. (5) We iteratively execute Steps (3) and (4)  until satisfying answers are achieved. We have at least 25 paraphrases: 5 short, 5 long, with the rest being medium length. Furthermore, we introduce a random split of our paraphrases into two distinct sets: a training set comprising 50\% of the entire dataset, and a test set constituting the remaining 50\%. The out-of-domain set encompasses all long and short paraphrases, aiming at simulating the situation where individuals seek to extract specific knowledge by inputting a concise  or exceptionally long query. 
We provide an example  in  Appendix \ref{Paratrex:details}.

\subsection{Evaluation} 
We evaluate the quality of ParaTrex using two automatic metrics and human evaluation.

\begin{figure}[ht]
\centering
\includegraphics[width=0.8\linewidth,height=0.5\linewidth]{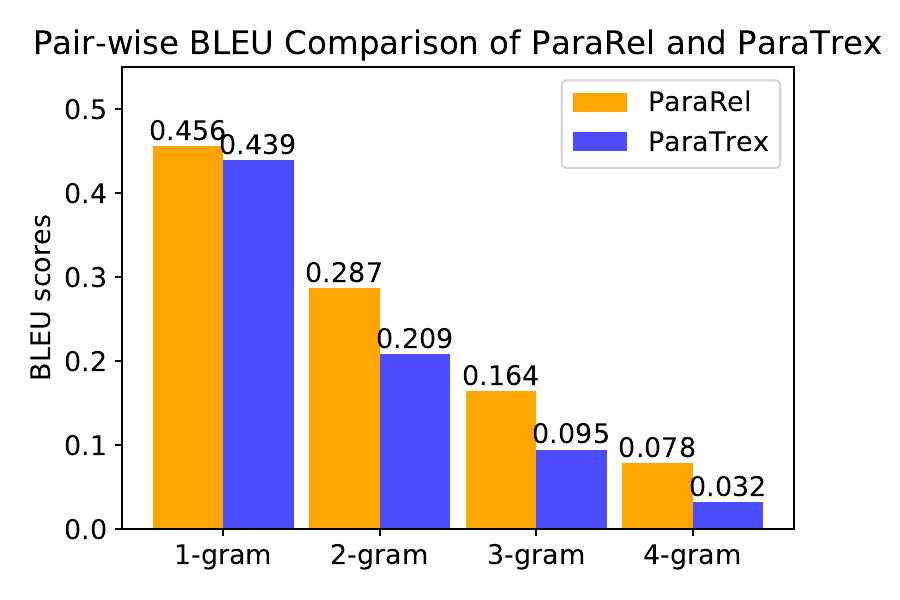}
\caption{Average pair-wise BLEU between all relations comparison with ParaRel. ParaTrex gets a consistently lower score than ParaRel, representing that the templates in ParaTrex are more lexically and syntactically diverse.}
\label{fig:bleu}
\end{figure}

\paragraph{Diversity} 
We test the lexical diversity by reporting the average pairwise BLEU scores of each relation. Specifically, all pair-wise permutations of $n$ templates for each relation are listed, resulting in $n(n-1)$ sentence pairs. Then the pair-wise $n$-gram BLEU score \citep{papineni2002bleu} was calculated to represent their diversity. The average score of the lower-order $n$-gram score captures lexical diversity and the average score of the higher-order $n$-gram score tends to capture the diversity of complex syntactic structures.  Fig \ref{fig:bleu} shows the trend over n-gram average pairwise BLEU scores of all relations. We find that the BLEU scores of ParaTrex perform consistently lower than ParaRel, which depicts that ParaTrex has a better lexical and syntactical diversity of generated sentences compared with the existing baseline datasets.

\begin{figure}[ht]
\centering
\includegraphics[width=1\linewidth,height=0.35\linewidth]{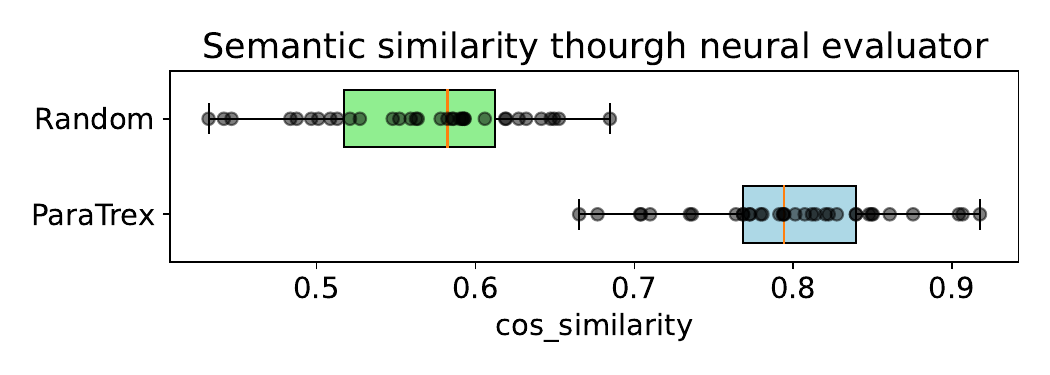}
\setlength{\abovecaptionskip}{-0.2cm}
\setlength{\belowcaptionskip}{-0.2cm}
\caption{The cosine similarity of the embedding between the grounding template and the paraphrased template. The boxplot shows the comparison between the random paraphrase sampled from other relations and the paraphrase in our dataset for 39 relations.}
\label{fig:cossimilarity}
\end{figure}

\paragraph{Quality} 
For automatic qualitative evaluation, we perform the current SoTA version \textit{paraphrase-multilingual-mpnet-base-v2} of Sentence-BERT \citep{reimers-2019-sentence-bert} on the Sentance-BERT leaderboard\footnote{https://www.sbert.net/docs/pre-trained\_models.html} to evaluate the semantic similarity between the paraphrase and the grounding prompt template provided in the LAMA dataset. We report the average cosine similarity upon all paraphrases for each relation in our dataset and show it in a boxplot (Fig \ref{fig:cossimilarity}). These results show that ParaTrex shares good semantic alignments with the grounding datasets except for two special cases where two relations get scores lower than $0.7$. This is because the grounding templates ``\textit{[X] plays [Y]}'' and ``\textit{[X] is located in [Y]}'' miss the information that [Y] refers to musical instruments and continents respectively. In contrast, this information is included in ParaTrex since it is provided in the description when constructing ParaTrex.

\paragraph{Human Agreement} 
Following \citet{elazar2021measuring}, we randomly picked 82 paraphrases in the ParaTrex dataset and 42 wrong paraphrases by sampling from the paraphrases of wrong relations.  We perform human evaluation by asking the evaluators to select candidates that are not the paraphrase of the given inputs. The participants need to pick out the wrong paraphrases. We consider the remaining answers as what they think to be the correct paraphrases of the given inputs. Two examples of questions are shown in Fig \ref{paratrex:he}. Results show that on average among 11 human judgments, human evaluators get 96.88\% accuracy in successfully identifying inaccurate paraphrases and a 92\% accuracy in selecting the true paraphrases provided by ParaTrex, which shows that our proposed datasets have a satisfying agreement with human beings, thus proving the favorable quality of our datasets.

\section{Methodology}
Based on the probability decomposition in Section \ref{prob:decompose}, we hypothesize that mitigating the misalignment between the tuning and pre-training objectives is the key to improving both the accuracy and consistency of models on unseen prompts. To this end, the core idea behind UniArk is to equalize the probability of biased parts through an additional loss and template augmentation. We discuss below  the  three main components of \textbf{UniArk}.

\smallskip \inlineSubsection{Adapters}
We use adapter-tuning~\citep{houlsby2019parameter} 
as it is better suited for debiasing settings \citep{kumar-etal-2023-parameter} and internal knowledge protections than other popular parameter-efficient fine-tuning methods. Moreover, we want to evaluate and thus fill in the vacancy of adapter-tuning on the factual knowledge extraction tasks. Note that for factual probing, it is common to tune a model for each relation. Due to the cost of storage when the relations scale up, we therefore do not choose full parameter fine-tuning as the basis of our framework. The basic idea is to insert an adapter into our base language models and freeze all other parameters. Specifically, for each output $\textbf{h}^n \in \mathbb{R}^d$ in the $n$-th transformer layer, our adapters perform the following transformation:
\begin{equation} 
    \mathbf{h}^{n+1} = \text{GELU}(\mathbf{h}^{n} \mathbf{W}_d)\mathbf{W}_u+\mathbf{h}
\end{equation}
where GELU \citep{hendrycks2016gaussian} is a non-linear activate function, $\mathbf{W}_d \in \mathbb{R}^{d\times k}$ and $\mathbf{W}_u \in \mathbb{R}^{k\times d}$ are two learnable parameter matrices in adapters. They are used for first down-projecting the hidden states into dimension $k<d$, and then projecting them back to $d$-dimension spaces, with $k$  a hyperparameter.

\smallskip
\inlineSubsection{Object likelihood Bias Mitigation}
As discussed in Section~\ref{sec:promptprefbias}, to mitigate the object likelihood bias, the output distribution should ideally satisfy: for all $o_i, o_j \in \mathcal{O}, s_i, s_j \in \mathcal{S}$ and $t_k \in \mathcal{T}$, we have that $ P(o_i|t_k) = P(o_j|t_k), P(s_i|t_k) = P(s_j|t_k)$. In other words, the retrieved likelihood distribution should be close to a uniform distribution from the subject-masked  and object-masked inputs. To this end, we introduce an addition max entropy loss $L_{me}$ weighted by hyperparameter $\lambda_{me}$ over subject-masked prompts and object-masked prompts. This loss maximizes the entropy over top retrieved candidates to encourage the model to assign equal probability to each relevant candidate. We perform an object filtering process  to remove stopwords like ``\textit{and}''. We choose to max the entropy of only the top $k$ words because, based on our empirical observation, they include most of the relevant candidates. Formally, given the output probability of object  $i: p(i), i=1,2,\ldots, k$ and the stopwords set $S$, the max entropy loss is:
\begin{equation} 
    \mathcal{L}_{me}=-\lambda_{me}\sum_{i=1,\ i \notin S}^{k}p(i)\text{log}_2(p(i))
\end{equation}
We note that unlike MeCoD \cite{wang-etal-2023-towards-alleviating}, our method introduces no additional parameter and focuses on equalizing the likelihood for all potential candidates while MeCoD performs neural object selecting and does contrastive learning over the selected objects. This suggests that our method is lighter than MeCoD. We also generalise MeCoD since we consider both subject-masked and object-masked prompts, guided by our objective decompositions.

\smallskip
\inlineSubsection{Template prior Bias Mitigation}
To alleviate the template prior bias, we propose a novel self-data augmentation method to mitigate the influence of $P(t_i)$ by weighted averaging them. We augment our raw data with prefixes ``\textit{It is true that}" and ``\textit{It is false that}" and encourage the model's self-consistency by a weighted average of their output distribution to make final predictions. Specifically, the output probability $P(o_i|s,t)$ for object candidate $i$ and the masked language model  (MLM) loss $L_{mlm}$ are calculated as:
\begin{align} 
    P(o_i|s,t) = \textit{softmax}(\sum_{t_j\in \mathcal{T}^*}w_jP(o_i|s,t_{j}))\\
    \mathcal{L}_{mlm} = -\sum_{i=1}^{n_{vocab}}y_i\text{log}P(o_i|s,t)
\end{align}
where $\mathcal{T}^*=\{t, t_{\text{true}}, t_{\text{false}}\}$ is the set of augmented prompt templates and the weight $\sum_{j}w_j=1$ is a hyperparameter balancing the weight for each template. Note that we set $w_{\text{true}}=-w_{\text{false}}$ since the prompts ``\textit{It is true that}'' and ``\textit{It is false that}''  give opposite predictions.

\section{Experiments}
\smallskip
\inlineSubsection{Dataset}
We use LAMA-TREx
\cite{petroni-etal-2019-language} as our main training dataset, with the same train-test splits following \citet{LIU2023}. This dataset comprises 41 relations and 29,500 testing triples. To test the generalising ability and consistency for different prompt templates, we test the model on two additional paraphrased datasets: our ParaTrex and ParaRel~\cite{elazar2021measuring}. In both datasets,
  N-M relations are omitted when measuring consistency since it can be hard to measure consistency among several correct answers. 25 relations remained after filtering those.

\smallskip
\inlineSubsection{Evaluation Metrics} 
We evaluate the performance of models on three aspects: quality of extraction, object likelihood bias, and template prior bias. (1) For measuring the quality, we evaluate the macro F1 score for each relation over LAMA (LM), ParaTrex (PT), and ParaRel (PR) to test its performance in in-domain settings and generalisation on out-of-domain prompt templates. (2) To test the bias from the object likelihood, we report the hitting rate of the candidates from the counterfactual subject-masked prompt (CT\_hit1). Additionally, we report the KL-divergence (KLD) between the subject-masked prompt and the original prompt to show the influence of the prompt template on the likelihood distribution of the final retrieved candidates. (3) For the template prior bias, we measure the consistency of paraphrases in both ParaTrex and ParaRel. Following~\citet{elazar2021measuring} and \citet{newman2021p1}, \emph{consistency} is calculated as the ratio of consistent predictions from different paraphrases with all the paraphrases permutations.
We also measure consistency between the unique raw prompt template from LAMA and the paraphrased templates. We refer to this consistency as \textit{raw\_cst} while consistency between all permutations as \textit{all\_cst}. The previous consistency measures do not consider strict factual accuracy. Thus, we  also  measure the consistency over factual correct predictions, called \textit{acc\_cst}. Formal definitions of \textit{raw\_cst}, \textit{all\_cst} and  \textit{acc\_cst} are in Appendix \ \ref{sec:consistency}. 

\smallskip
\inlineSubsection{Baselines} We split our experiments into two settings: soft and manual prompts. In the former setting, we choose P-tuning \citep{LIU2023}, a popular prompt-tuning method in knowledge probing tasks, and  the SoTA MeCoD \citep{wang-etal-2023-towards-alleviating} as baselines. We compare them with the adapter-tuning to explore its performance. Note that we cannot measure the consistency over paraphrases here since the prompt template is learned through training. For the manual prompt setting, we take the manual prompt without tuning (LAMA) and adapter-tuning as baselines. Additionally, we re-implement MeCoD as MeCoD (OI) through adapter-tuning as it is originally based on P-tuning.  Appendix \ref{sec:trainingdetails} provides more training details.

\smallskip
\inlineSubsection{Significance Test}
To test the significance of any improvements or deterioration, we perform the following tests between our UniArk and the adapters baseline: (1) Paired T-Test and Wilcoxon Sign Test for a fixed seed among results across all relations and (2)  T-test among the averaged values of all relations after running UniArk with three different seeds. See detailed results in the Appendix \ref{sec:sg_test}.

\begin{table*}[ht]
\centering
\small
\begin{tabular}{lcccccccccc}
\toprule
  \multicolumn{1}{c}{\multirow{4}{*}{Method}} &
  \multicolumn{5}{c}{BERT-Large} &
  \multicolumn{5}{c}{RoBERTa-Large}
  \\
  \cmidrule(lr){2-11}
  &
  \multicolumn{2}{c}{OOD} &
  \multicolumn{1}{c}{ID} &
  \multicolumn{2}{c}{OL Bias} &
  \multicolumn{2}{c}{OOD} &
  \multicolumn{1}{c}{ID} &
  \multicolumn{2}{c}{OL Bias} 
  \\
  \cmidrule(lr){2-6}\cmidrule(lr){7-11} 
  &
  PT\_F1 &
  PR\_F1 &
  LM\_F1 &
  CT\_hit1 &
  KLD &
  PT\_F1 &
  PR\_F1 &
  LM\_F1 &
  CT\_hit1 &
  KLD\\ \midrule
  P-tuning &
  \multicolumn{2}{c}{\multirow{3}{*}{-}} &
  29.94 &
  15.91 &
  \phantom{0}3.34 &
  \multicolumn{2}{c}{\multirow{3}{*}{-}} &
  19.36 &
  17.13 &
  \phantom{0}2.06 \\
  +MeCoD &
  \multicolumn{2}{c}{} &
  29.33 &
  \phantom{0}1.02 &
  \textit{\phantom{0}8.48} &
  \multicolumn{2}{c}{} &
  23.13&
  \phantom{0}5.67&
  \phantom{0}5.39\\
 +Adapters &
  \multicolumn{2}{c}{} &
  31.21 &
  14.00 &
  \phantom{0}3.40 &
  \multicolumn{2}{c}{} &
  27.70 &
  14.72 &
  \phantom{0}3.47\\ \midrule
  LAMA &
  14.21 &
  16.00 &
  20.68 &
  \phantom{0}4.19 &
  \phantom{0}3.57 &
  \phantom{0}8.34 &
  \phantom{0}9.19 &
  12.37 &
  \phantom{0}5.23 &
  \phantom{0}1.83 \\
  Adapters &
  24.69 &
  27.34 &
  \textit{32.10} &
  12.77 &
  \phantom{0}5.54 &
  \textit{22.12} &
  \textit{23.78} &
  \textbf{29.74} &
  16.88 &
  \phantom{0}3.40
  \\
  +MeCoD (OI) &
  \textit{25.64} &
  \textit{27.58} &
  31.79 &
  \textit{\phantom{0}0.13} &
  \phantom{0}7.31 &
  21.97 &
  23.34 &
  28.72 &
  \textit{\phantom{0}5.00} &
  \textit{\phantom{0}6.13}
  \\
  \textbf{+UniArk} &
  \textbf{{\ul27.99}} &
  \textbf{{\ul28.48}} &
  {\textbf{32.14}} &
  {\textbf{{\ul \phantom{0}0.04}}} &
  \textbf{{\ul11.66}} &
  \textbf{{\ul23.68}} &
  \textbf{{\ul24.70}} &
  \textit{29.29} &
  \textbf{{\ul\phantom{0}3.65}} &
  \textbf{{\ul10.24}}
  \\
  \midrule
  Fine-tune &
  28.50 &
  29.27 &
  30.85 &
  13.11 &
  \phantom{0}8.07 &
  25.05 &
  25.53 &
  27.85 &
  12.23 &
  \phantom{0}6.11
  \\ \bottomrule
\end{tabular}
\setlength{\belowcaptionskip}{-0.2cm}
\caption{
Main results for out-of-main (OOD), in-domain (ID) performance, and object likelihood bias (OL Bias) on LAMA (averaged over all relations). The underlines represent the significance after three significance tests.}
\label{Tabel3}
\end{table*}

\begin{table}[ht]
\centering\small
\setlength{\tabcolsep}{0.6ex}
\begin{tabular}{clcccccc}
\toprule
\multirow{2}{*}{Model} &
\multicolumn{1}{c}{\multirow{2}{*}{Method}} &
  \multicolumn{3}{c}{ParaTrex} &
  \multicolumn{3}{c}{ParaRel} \\
\cmidrule(lr){3-5}\cmidrule(lr){6-8} 
& &  \multicolumn{1}{r}{raw} &
  \multicolumn{1}{c}{all} &
  \multicolumn{1}{c}{acc} &
  \multicolumn{1}{r}{raw} &
  \multicolumn{1}{c}{all} &
  \multicolumn{1}{c}{acc} \\
\midrule
\multicolumn{1}{c}{\multirow{5}{*}{\makecell{Roberta\\-large}}} &
  LAMA &
  23.9 &
  20.6 &
  \multicolumn{1}{c}{\phantom{0}6.9} &
  33.0 &
  28.3 &
  10.4 \\
 &
  Adapters &
  61.9 &
  55.2 &
  \multicolumn{1}{c}{34.1} &
  66.9 &
  60.4 &
  37.3 \\
 &
  + MeCoD (OI) &
  61.7 &
  54.8 &
  \multicolumn{1}{c}{34.6} &
  67.9 &
  61.2 &
  38.1 \\
 &
  \textbf{+ UniArk} &
  \textbf{{\ul63.8}} &
  \textbf{{\ul59.0}} &
  \multicolumn{1}{c}{\textbf{{\ul36.2}}} &
  \textbf{{\ul69.1}} &
  \textbf{{\ul63.4}} &
  \textbf{{\ul38.5}} \\
\midrule
\multicolumn{1}{c}{\multirow{5}{*}{\makecell{BERT\\-large}}} &
  LAMA &
  33.6 &
  28.3 &
  \multicolumn{1}{c}{15.8} &
  54.9 &
  46.6 &
  25.0 \\
 &
  Adapters &
  60.9 &
  53.4 &
  \multicolumn{1}{c}{39.1} &
  72.1 &
  65.2 &
  45.8 \\
 &
  + MeCoD (OI) &
  63.4 &
  56.5 &
  \multicolumn{1}{c}{41.2} &
  73.5 &
  67.3 &
  47.2 \\
 &
  \textbf{+ UniArk} &
  \textbf{{\ul69.1}} &
  \textbf{{\ul62.9}} &
  \multicolumn{1}{c}{\textbf{{\ul 44.7}}} &
  \textbf{{\ul76.7}} &
  \textbf{{\ul71.3}} &
  \textbf{{\ul49.4}} \\

\bottomrule
\end{tabular}
\setlength{\belowcaptionskip}{-0.2cm}
\caption{Main results for template prior bias (TP bias) measured by consistency on ParaTrex and ParaRel. Significantly improved results are underlined.}
\label{Table4}
\end{table}

\subsection{Quantitative Results}
Table~\ref{Tabel3} presents results for knowledge retrieval quality together with object likelihood bias on BERT-large \cite{Devlin2019bert} and RoBERTa-large \cite{liu2019roberta}. Table~\ref{Table4} shows results for template prior bias. The best value is marked in bold and the second best value is marked in italics.

\smallskip
\inlineSubsection{Main Results}
For probing quality, we find that with the appropriate tuning methods, models with manual prompts outperform those with soft prompting. This shows the necessity of tuning parameters within the models rather than within the input embeddings. Among all vanilla tuning methods, Adapters demonstrate a remarkable capability for in-domain knowledge and object likelihood bias. They outperform fine-tuning over $0.01$ ($4\%$) on the in-domain F1-score, with also less object likelihood bias than P-tuning and fine-tuning. However, it is still shown to be under severe biases and performs poorly on the out-of-domain prompts. With our proposed framework UniArk for mitigating both biased objectives, we significantly improve the generalisation ability to probe knowledge on unseen prompts. Various significance tests prove the improvements in the out-of-domain generalisations and two bias mitigations over adapters and MeCoD baselines. The in-domain quality is also shown not harmed. Indeed, UniArk outperforms the current SoTA MeCoD in both in-domain and out-of-domain prompt templates.

\smallskip
\inlineSubsection{Adapters versus Other Tuning Methods}
To better understand the capabilities of the adapter-tuning method on factual knowledge extraction, we compare it with manual prompts (LAMA), P-tuning (PT), and fine-tuning (FT). We do not consider other parameter-efficient fine-tuning methods, such as prefix-tuning \cite{li-liang-2021-prefix}, since they are shown to be less powerful than P-tuning 
\cite{LIU2023,wang-etal-2023-towards-alleviating}. Table \ref{Tabel3} shows that the adapter-tuning performs consistently better than all other parameter-efficient fine-tuning methods in the F1 score when tuning on the in-domain settings. This strongly suggests that tuning methods such as adapters, which modify the inner transformer layers instead of only embedding layers without changing the initial parameters, may do better in extracting the knowledge hard encoded within the parameters in LMs. However, there exists a substantial gap in  performance  between in-domain and out-of-domain settings. Indeed, we observe a big gap in F1 scores, suggesting that those parameter-efficient tuning methods tend to be biased on the given prompt template.

\smallskip
\inlineSubsection{Bias Mitigation and Quality Improvements}
As Table \ref{Tabel3} shows, with our proposed framework UniArk, both object likelihood bias and prompt prior bias are effectively mitigated. The counterfactual hitting rate drops to nearly 0. This means the model can no longer guess the correct answers given only templates. The sharp rise of KL-divergence also indicates that the model tends to predict a distribution diverging substantially from the object likelihood under prompt templates. Both metrics show that the model is no longer influenced by the object likelihood. Additionally, in Table \ref{Table4}, the consistency over all paraphrased datasets increases significantly, showing the effectiveness of our prior bias mitigation module. At the same time, we respectively observe improvements of  7\% (22.12 to 23.68), 4\% (23.78 to 24.7), and 13\% (24.69 to 27.99), 4\% (27.34 to 28.48) of out-of-domain F1 score in UniArk compared with the adapters baseline for RoBERTa and BERT on ParaTrex and ParaRel. This validates our hypothesis that mitigating the two decomposed bias terms helps generalisation to unseen prompts. 
Besides, we report the consistency after removing semantic overlapped relations stated in \cite{hagstrom-etal-2023-effect} in Appendix \ref{appendix:semantic overlap}, which follows a similar consistency trend, suggesting that the overlap does not influence the main result. We also provide a scaling study in Appendix \ \ref{scalestudy}, where we show that UniArk has significant improvement on both base and larger models. 

\subsection{Ablation Studies}
\begin{table}[t]
\centering
\small
\setlength{\tabcolsep}{0.7ex}
\begin{tabular}{lcccccc}
\toprule
        \multicolumn{1}{c}{\multirow{2}{*}{Method}} & 
        \multicolumn{2}{c}{Quality} & \multicolumn{2}{c}{OL Bias}     & \multicolumn{2}{c}{TP Bias} \\ \cmidrule(lr){2-7}
  & PT             & PR    & CT\_hit1        & KLD            & PT         & PR         \\
\midrule
UniArk & \textbf{28.0} & \textbf{28.5} & \textbf{\phantom{0}0.0} & 11.7          & \textbf{62.9}      & \textbf{71.3}      \\
w/o ME  & 26.9          & 28.4 & 13.2          & \phantom{0}5.5           & 60.8      & 70.5      \\
w/o Aug & 25.3          & 27.3 & \textbf{\phantom{0}0.0} & \textbf{12.3} & 56.0      & 66.3      \\
w/o ME \& Aug        & 24.7 & 27.3          & 16.9 &  \phantom{0}3.4  & 55.2          & 60.4          \\ \bottomrule
\end{tabular}
\setlength{\belowcaptionskip}{-0.4cm}
\caption{Ablation study on BERT, we report F1 score for extraction quality; and all\_consistency for template prior bias on ParaTrex (PT) and ParaRel (PR)}
\label{ablation}
\end{table}

We take adapter-tuning as a baseline and perform ablation studies  to locate the source of performance improvement. The results  in Table \ref{ablation} demonstrate that our max entropy (ME) module plays a prominent  role in relieving object likelihood bias while our self-augmenting (Aug) module makes the main contribution to mitigating prompt preference bias. Both modules increase the F1 scores of extraction quality, showing the help of bias mitigation for improving the out-of-domain generalisation. 

We emphasize that our ME module contributes to improving consistency and our Aug module brings an improvement on the prompt preference bias as well. This exhibits a synergizing effect of both modules on mitigating both biases, further highlighting the necessity of simultaneously alleviating biases within a unified framework. This effect is probably because, as we equalize the object likelihood over templates, the model is forced to treat the prompt templates as the same, which also weakens the favor of specific templates and thus increases the consistency over unseen prompts. Meanwhile, augmenting the templates forces the model to estimate the object likelihood over various cases, and averaging this likelihood distribution contributes to a more unbiased object likelihood.

\subsection{Qualitative Case Studies}
To better understand how mitigating the studied biases helps to improve the knowledge extraction results, we perform two specific case studies on randomly selected cases. A detailed analysis can be found in Appendix \ref{sec:quality study}. Here we give one example from each biased objective mitigation. For  template prior bias (Table \ref{case-pv}), although both UniArk and adapter-tuning make a correct prediction ``\textit{Finnish}'' on the question ``\textit{The official language of Vesanto is [mask]}'', the answers of adapters may turn to some pronoun such as  ``\textit{It}'' when the templates changed. UniArk relieves these kinds of errors with the augmented inputs and drops the predictions for ``It'' from 861 (7.4\%) times to 140 (1.2\%) times among all predictions in this relation according to our statistics. For object likelihood bias (Table \ref{case-pp}), when it comes to the question ``\textit{The official language of Sorengo is [mask]}'', the golden truth should be ``Italian''. However, traditional probing gives ``\textit{Portuguese}'' as the answer and we found that the rank 2, and rank 3 predictions ``\textit{English}'' and ``\textit{Spanish}'' appears in the prediction from the top and third predictions from subject-masked prompt, suggesting that the prediction of a traditional model may be influenced by this object likelihood. In contrast, UniArk, who provides the correct answers, is not influenced by this object ``\textit{English}'' since the subject-masked likelihood is uniformly distributed.

\section{Further Analysis}

\smallskip
\inlineSubsection{Using Paraphrased Data for Training}
To simulate real applications in which paraphrased data is lacking (and for a fair comparison), UniArk is tuned on a single prompt template provided in the LAMA dataset.  We try to investigate the following question: What if we use the part of paraphrased data for training? We added a new module called ``PARA'' following \cite{elazar2021measuring}, where an additional KL-Divergence loss between the prediction distribution from the LAMA template and the paraphrased template is added. We randomly select 1, 2, and 5 new paraphrased templates to perform experiments. From Table \ref{paratrain}, only a subtle improvement can be witnessed after adding new paraphrases to UniArk for training and these improvements also do not scale up with more given paraphrases. This indicates that our proposed self-data augmentation, with no paraphrases, is as powerful as training on paraphrases under current frameworks. This result also suggests a potential research direction for incorporating paraphrased data both efficiently and effectively during training.

\begin{table}[t]
\centering
\small
\setlength{\tabcolsep}{1ex}
\begin{tabular}{lcccccc}
\toprule
\multicolumn{1}{c}{\multirow{2}{*}{Method}} & 
\multicolumn{2}{c}{Quality (f1)} & \multicolumn{2}{c}{OL Bias}     & \multicolumn{2}{c}{TP Bias (cst)} \\ 
\cmidrule(lr){2-7}
  & PT             & PR    & CT\_hit1        & KLD            & PT         & PR         \\
\midrule
UniArk 
& 28.0 
& 28.5 
& 0.0 
& 11.7          
& 62.9      
& 71.3      
\\
+para 1 
& 28.1          
& 28.6 
& 0.0          
& 11.6          
& 63.3     
& 71.8      
\\
+para 2 
& 28.3          
& 28.9 
& 0.0
& 11.5 
& 63.3      
& 71.9      
\\
+para 5 
& 28.1 
& 28.6          
& 0.0 
& 11.6  
& 63.2          
& 71.8          
\\ \bottomrule
\end{tabular}
\setlength{\belowcaptionskip}{-0.4cm}
\caption{Results using paraphrased data for training. PT and PR refer to Paratrex and ParaRel respectively}
\label{paratrain}
\end{table}

\smallskip
\inlineSubsection{Error Analysis}
To have a comprehensive understanding of the existing errors in our factual probing framework, we conducted a random sampling of 50 incorrect predictions within the relation P37 ``\textit{Official\_Languages}'' We categorized these errors, documenting the findings in Appendix \ref{error_ans}. In summary, we find that LMs still do not have a comprehensive knowledge of specific cities such as Azad Kashmir. They also make mistakes in predicting pronouns like ``\textit{It}'' (4 cases), and in spelling (2 cases). Besides, we found 21 (42\%) cases where the model makes a feasible answer among several correct answers but is treated wrong because only one of the labels is provided, e.g. Finnish for Turku, suggesting that we may underestimate the knowledge stored in LMs via current metrics.

\section{Related Work}

\paragraph{Factual Knowledge Extraction}
There are several works on how to treat LMs as knowledge bases and extract factual knowledge from the weights of an LM. \citet{petroni-etal-2019-language} is one of the seminal works on this and also introduces the LAMA benchmark for extracting factual knowledge from LMs. To access the knowledge, \citet{Li2022TaskspecificPA} applies further pre-training (fine-tuning) on LMs. \citet{liu2023pre} suggests that manual prompts offer a promising avenue for directly accessing this knowledge without the need for extra fine-tuning.  
Recent works look at soft prompts with continuous learnable prompts. \citet{LIU2023} proposes P-tuning, making all tokens within prompt templates as learnable soft prompts and showing similar scaling results on larger language models. However, we observe that  adapter-tuning \cite{houlsby2019parameter} has not been applied to this task so far.
In this paper, we show that   adapter-tuning can be   a promising and robust way of factual knowledge extraction.

\smallskip
\inlineSubsection{Bias study}
\citet{cao-etal-2022-prompt} and \citet{elazar2021measuring} argue that there exist severe risks and biases under prompt-based knowledge extraction. Therefore, \citet{newman2021p1} attempt to increase the consistency by asserting a single multiple-layer perception after embedding layers. 
\citet{wang-etal-2023-towards-alleviating} propose the contrastive learning-based framework MeCoD for mitigating the bias. In this paper, we position and decompose the object likelihood bias and template prior bias and propose a unified framework for mitigating them, which is a more general case compared with previous studies.
As a concurrent work, \citet{wang2023assessing} propose a new metric and a dataset for measuring the reliability of factual probing.

\smallskip
\inlineSubsection{Model Editing}
As parametric knowledge from LMs might be outdated, there is a recent trend in editing LMs. \citet{meng2022locating,meng2022memit} proposed batch editing for models. \citet{HLLP2023} design explicit and
implicit multi-editor models to learn diverse editing strategies. \citet{huangtransformer}  addressed the problem of Sequential Model Editing. \citet{HLTW+2023} proposed a plug-and-play  retrieval augmented framework. \citet{tan2023massive} tried to edit massive knowledge via meta-learning.

\section{Conclusion}
In this paper, we revisit the factual probing objectives under a probabilistic view and point out the misalignment between the pre-training and fine-tuning objectives. This motivates our hypothesis that mitigating both template prior and object likelihood bias may improve the generalisability of knowledge-probing models. We  introduce ParaTrex, a large and high-quality dataset for measuring the generalisability, and propose a parameter-free method to validate this hypothesis. Experiments show the superiority of our framework and a synergizing effect is found by alleviating both biases, proving the necessity of a unified framework towards a generalised factual knowledge extraction.

\section*{Limitations}
We identify the following two limitations  related to the methodology and base models and one limitation for the dataset ParaTrex. First, in our verbalization bias mitigating module, we  perform a naive average between the self-augmenting inputs and the original inputs, following our objective decomposition parts.  Although it works effectively, it would be interesting to investigate other   methods. Second, the prompt template in LAMA and ParaTrex/ParaRel datasets is designed for  masked language modeling instead of  next token prediction. We made a scaling study on  encoder-only models to show the scalability of our methods, it would be interesting to also construct corresponding datasets for decoder-only large language models and perform experiments on them. We leave this for future work. For ParaTrex, we mitigated but did not completely solve relations containing unidiomatic templates \cite{hagstrom-etal-2023-effect}. For example, in some cases, the models are more likely to give correct predictions over template ``[X] works as a/an [Y]'' compared with ``[X] works as [Y]''.

\section*{Ethics Statement}
During the construction of the paraphrased dataset ParaTrex,  we did not generate any data that is harmful to society and humans, nor include any private personal information within the dataset.

\section*{Acknowledgments}
This work originated from a dissertation at the University of Edinburgh. We would like to acknowledge Chenmien Tan for his thoughtful feedback and suggestions throughout the writing process of this work. In addition, we thank Hanxu Hu, Simon Yu, and Yifu Qiu for giving valuable suggestions on a draft of this work and thank the anonymous reviewers for their advice to further clarify the ParaTrex quality and the novelty of this work.

\bibliography{custom}

\begin{thebibliography}{31}
\expandafter\ifx\csname natexlab\endcsname\relax\def\natexlab#1{#1}\fi

\bibitem[{Cao et~al.(2022)Cao, Lin, Han, Liu, and Sun}]{cao-etal-2022-prompt}
Boxi Cao, Hongyu Lin, Xianpei Han, Fangchao Liu, and Le~Sun. 2022.
\newblock \href {https://doi.org/10.18653/v1/2022.acl-long.398} {Can prompt probe pretrained language models? understanding the invisible risks from a causal view}.
\newblock In \emph{Proceedings of the 60th Annual Meeting of the Association for Computational Linguistics (Volume 1: Long Papers)}.

\bibitem[{Cao et~al.(2021)Cao, Lin, Han, Sun, Yan, Liao, Xue, and Xu}]{cao-etal-2021-knowledgeable}
Boxi Cao, Hongyu Lin, Xianpei Han, Le~Sun, Lingyong Yan, Meng Liao, Tong Xue, and Jin Xu. 2021.
\newblock \href {https://doi.org/10.18653/v1/2021.acl-long.146} {Knowledgeable or educated guess? revisiting language models as knowledge bases}.
\newblock In \emph{Proceedings of the 59th Annual Meeting of the Association for Computational Linguistics and the 11th International Joint Conference on Natural Language Processing (Volume 1: Long Papers)}.

\bibitem[{Devlin et~al.(2019)Devlin, Chang, Lee, and Toutanova}]{Devlin2019bert}
Jacob Devlin, Ming{-}Wei Chang, Kenton Lee, and Kristina Toutanova. 2019.
\newblock \href {https://doi.org/10.18653/v1/n19-1423} {{BERT:} pre-training of deep bidirectional transformers for language understanding}.
\newblock In \emph{Proceedings of the 2019 Conference of the North American Chapter of the Association for Computational Linguistics: Human Language Technologies, {NAACL-HLT} 2019, Minneapolis, MN, USA, June 2-7, 2019, Volume 1 (Long and Short Papers)}.

\bibitem[{Elazar et~al.(2021)Elazar, Kassner, Ravfogel, Ravichander, Hovy, Sch{\"u}tze, and Goldberg}]{elazar2021measuring}
Yanai Elazar, Nora Kassner, Shauli Ravfogel, Abhilasha Ravichander, Eduard Hovy, Hinrich Sch{\"u}tze, and Yoav Goldberg. 2021.
\newblock \href {https://direct.mit.edu/tacl/article/doi/10.1162/tacl_a_00410/107384/Measuring-and-Improving-Consistency-in-Pretrained} {Measuring and improving consistency in pretrained language models}.
\newblock \emph{Transactions of the Association for Computational Linguistics}.

\bibitem[{Hagstr{\"o}m et~al.(2023)Hagstr{\"o}m, Saynova, Norlund, Johansson, and Johansson}]{hagstrom-etal-2023-effect}
Lovisa Hagstr{\"o}m, Denitsa Saynova, Tobias Norlund, Moa Johansson, and Richard Johansson. 2023.
\newblock \href {https://doi.org/10.18653/v1/2023.emnlp-main.332} {The effect of scaling, retrieval augmentation and form on the factual consistency of language models}.
\newblock In \emph{Proceedings of the 2023 Conference on Empirical Methods in Natural Language Processing}.

\bibitem[{Han et~al.(2023{\natexlab{a}})Han, Li, Li, and Pan}]{HLLP2023}
Xiaoqi Han, Ru~Li, Xiaoli Li, and Jeff~Z. Pan. 2023{\natexlab{a}}.
\newblock \href {https://doi.org/https://doi.org/10.1016/j.knosys.2023.110826} {{A Divide and Conquer Framework for Knowledge Editing}}.
\newblock \emph{Knowledge Based Systems}, 279.

\bibitem[{Han et~al.(2023{\natexlab{b}})Han, Li, Tan, Yuanlong, Chai, and Pan}]{HLTW+2023}
Xiaoqi Han, Ru~Li, Hongye Tan, Wang Yuanlong, Qinghua Chai, and Jeff~Z. Pan. 2023{\natexlab{b}}.
\newblock \href {https://aclanthology.org/2023.findings-emnlp.749/} {{Improving Sequential Model Editing with Fact Retrieval}}.
\newblock In \emph{Findings of the Association for Computational Linguistics (EMNLP 2023)}.

\bibitem[{Hendrycks and Gimpel(2016)}]{hendrycks2016gaussian}
Dan Hendrycks and Kevin Gimpel. 2016.
\newblock \href {https://arxiv.org/abs/1606.08415} {Gaussian error linear units (gelus)}.
\newblock \emph{arXiv preprint arXiv:1606.08415}.

\bibitem[{Houlsby et~al.(2019)Houlsby, Giurgiu, Jastrzebski, Morrone, De~Laroussilhe, Gesmundo, Attariyan, and Gelly}]{houlsby2019parameter}
Neil Houlsby, Andrei Giurgiu, Stanislaw Jastrzebski, Bruna Morrone, Quentin De~Laroussilhe, Andrea Gesmundo, Mona Attariyan, and Sylvain Gelly. 2019.
\newblock \href {https://proceedings.mlr.press/v97/houlsby19a} {Parameter-efficient transfer learning for nlp}.
\newblock In \emph{International Conference on Machine Learning}. PMLR.

\bibitem[{Huang et~al.(2023)Huang, Shen, Zhang, Zhou, Rong, and Xiong}]{huangtransformer}
Zeyu Huang, Yikang Shen, Xiaofeng Zhang, Jie Zhou, Wenge Rong, and Zhang Xiong. 2023.
\newblock \href {https://openreview.net/pdf?id=4oYUGeGBPm} {Transformer-patcher: One mistake worth one neuron}.
\newblock In \emph{The Eleventh International Conference on Learning Representations}.

\bibitem[{Kumar et~al.(2023)Kumar, Lesota, Zerveas, Cohen, Eickhoff, Schedl, and Rekabsaz}]{kumar-etal-2023-parameter}
Deepak Kumar, Oleg Lesota, George Zerveas, Daniel Cohen, Carsten Eickhoff, Markus Schedl, and Navid Rekabsaz. 2023.
\newblock \href {https://doi.org/10.18653/v1/2023.eacl-main.201} {Parameter-efficient modularised bias mitigation via {A}dapter{F}usion}.
\newblock In \emph{Proceedings of the 17th Conference of the European Chapter of the Association for Computational Linguistics}.

\bibitem[{Lauscher et~al.(2021)Lauscher, Lueken, and Glava{\v{s}}}]{lauscher-etal-2021-sustainable-modular}
Anne Lauscher, Tobias Lueken, and Goran Glava{\v{s}}. 2021.
\newblock \href {https://doi.org/10.18653/v1/2021.findings-emnlp.411} {Sustainable modular debiasing of language models}.
\newblock In \emph{Findings of the Association for Computational Linguistics: EMNLP 2021}.

\bibitem[{Li et~al.(2022)Li, Huang, Papasarantopoulos, Vougiouklis, and Pan}]{Li2022TaskspecificPA}
Tianyi Li, Wenyu Huang, Nikos Papasarantopoulos, Pavlos Vougiouklis, and Jeff~Z. Pan. 2022.
\newblock \href {https://api.semanticscholar.org/CorpusID:251881464} {Task-specific pre-training and prompt decomposition for knowledge graph population with language models}.
\newblock \emph{ArXiv}, abs/2208.12539.

\bibitem[{Li and Liang(2021)}]{li-liang-2021-prefix}
Xiang~Lisa Li and Percy Liang. 2021.
\newblock \href {https://doi.org/10.18653/v1/2021.acl-long.353} {Prefix-tuning: Optimizing continuous prompts for generation}.
\newblock In \emph{Proceedings of the 59th Annual Meeting of the Association for Computational Linguistics and the 11th International Joint Conference on Natural Language Processing (Volume 1: Long Papers)}.

\bibitem[{Liu et~al.(2023{\natexlab{a}})Liu, Yuan, Fu, Jiang, Hayashi, and Neubig}]{liu2023pre}
Pengfei Liu, Weizhe Yuan, Jinlan Fu, Zhengbao Jiang, Hiroaki Hayashi, and Graham Neubig. 2023{\natexlab{a}}.
\newblock \href {https://dl.acm.org/doi/full/10.1145/3560815} {Pre-train, prompt, and predict: A systematic survey of prompting methods in natural language processing}.
\newblock \emph{ACM Computing Surveys}, 55(9).

\bibitem[{Liu et~al.(2023{\natexlab{b}})Liu, Zheng, Du, Ding, Qian, Yang, and Tang}]{LIU2023}
Xiao Liu, Yanan Zheng, Zhengxiao Du, Ming Ding, Yujie Qian, Zhilin Yang, and Jie Tang. 2023{\natexlab{b}}.
\newblock \href {https://doi.org/https://doi.org/10.1016/j.aiopen.2023.08.012} {Gpt understands, too}.
\newblock \emph{AI Open}.

\bibitem[{Liu et~al.(2019)Liu, Ott, Goyal, Du, Joshi, Chen, Levy, Lewis, Zettlemoyer, and Stoyanov}]{liu2019roberta}
Yinhan Liu, Myle Ott, Naman Goyal, Jingfei Du, Mandar Joshi, Danqi Chen, Omer Levy, Mike Lewis, Luke Zettlemoyer, and Veselin Stoyanov. 2019.
\newblock \href {https://arxiv.org/abs/1907.11692} {Roberta: A robustly optimized bert pretraining approach}.
\newblock \emph{arXiv preprint arXiv:1907.11692}.

\bibitem[{Meng et~al.(2022)Meng, Bau, Andonian, and Belinkov}]{meng2022locating}
Kevin Meng, David Bau, Alex Andonian, and Yonatan Belinkov. 2022.
\newblock \href {https://proceedings.neurips.cc/paper_files/paper/2022/hash/6f1d43d5a82a37e89b0665b33bf3a182-Abstract-Conference.html} {Locating and editing factual associations in gpt}.
\newblock \emph{Advances in Neural Information Processing Systems}, 35:17359--17372.

\bibitem[{Meng et~al.(2023)Meng, Sen~Sharma, Andonian, Belinkov, and Bau}]{meng2022memit}
Kevin Meng, Arnab Sen~Sharma, Alex Andonian, Yonatan Belinkov, and David Bau. 2023.
\newblock \href {https://openreview.net/forum?id=MkbcAHIYgyS} {Mass editing memory in a transformer}.
\newblock In \emph{International Conference on Machine Learning}.

\bibitem[{Newman et~al.(2022)Newman, Choubey, and Rajani}]{newman2021p1}
Benjamin Newman, Prafulla~Kumar Choubey, and Nazneen Rajani. 2022.
\newblock \href {https://openreview.net/forum?id=DhzIU48OcZh} {P-adapters: Robustly extracting factual information from language models with diverse prompts}.
\newblock In \emph{The Tenth International Conference on Learning Representations, {ICLR}}.

\bibitem[{Pan et~al.(2017{\natexlab{a}})Pan, Vetere, Gomez-Perez, and Wu}]{PVGW2017}
J.~Z. Pan, G.~Vetere, J.M. Gomez-Perez, and H.~Wu, editors. 2017{\natexlab{a}}.
\newblock \href {https://link.springer.com/book/10.1007/978-3-319-45654-6} {\emph{{Exploiting Linked Data and Knowledge Graphs for Large Organisations}}}.
\newblock Springer.

\bibitem[{Pan et~al.(2023)Pan, Razniewski, Kalo, Singhania, Chen, Dietze, Jabeen, Omeliyanenko, Zhang, Lissandrini, Biswas, de~Melo, Bonifati, Vakaj, Dragoni, , and Graux}]{PRKS+2023}
Jeff~Z. Pan, Simon Razniewski, Jan-Christoph Kalo, Sneha Singhania, Jiaoyan Chen, Stefan Dietze, Hajira Jabeen, Janna Omeliyanenko, Wen Zhang, Matteo Lissandrini, Russa Biswas, Gerard de~Melo, Angela Bonifati, Edlira Vakaj, Mauro Dragoni, , and Damien Graux. 2023.
\newblock \href {https://drops.dagstuhl.de/entities/document/10.4230/TGDK.1.1.2} {{Large Language Models and Knowledge Graphs: Opportunities and Challenges}}.
\newblock \emph{Special Issue on Trends in Graph Data and Knowledge. Transactions on Graph Data and Knowledge (TGDK)}, 1:1--38.

\bibitem[{Pan et~al.(2017{\natexlab{b}})Pan, Calvanese, Eiter, Horrocks, Kifer, Lin, and Zhao}]{Pan2017b}
J.Z. Pan, D.~Calvanese, T.~Eiter, I.~Horrocks, M.~Kifer, F.~Lin, and Y.~Zhao. 2017{\natexlab{b}}.
\newblock \href {https://link.springer.com/book/10.1007/978-3-319-49493-7} {\emph{{Reasoning Web: Logical Foundation of Knowledge Graph Construction and Querying Answering}}}.
\newblock Springer.

\bibitem[{Papineni et~al.(2002)Papineni, Roukos, Ward, and Zhu}]{papineni2002bleu}
Kishore Papineni, Salim Roukos, Todd Ward, and Wei-Jing Zhu. 2002.
\newblock \href {https://doi.org/10.3115/1073083.1073135} {{B}leu: a method for automatic evaluation of machine translation}.
\newblock In \emph{Proceedings of the 40th Annual Meeting of the Association for Computational Linguistics}.

\bibitem[{Petroni et~al.(2019)Petroni, Rockt{\"a}schel, Riedel, Lewis, Bakhtin, Wu, and Miller}]{petroni-etal-2019-language}
Fabio Petroni, Tim Rockt{\"a}schel, Sebastian Riedel, Patrick Lewis, Anton Bakhtin, Yuxiang Wu, and Alexander Miller. 2019.
\newblock \href {https://doi.org/10.18653/v1/D19-1250} {Language models as knowledge bases?}
\newblock In \emph{Proceedings of the 2019 Conference on Empirical Methods in Natural Language Processing and the 9th International Joint Conference on Natural Language Processing (EMNLP-IJCNLP)}.

\bibitem[{Reimers and Gurevych(2019)}]{reimers-2019-sentence-bert}
Nils Reimers and Iryna Gurevych. 2019.
\newblock \href {http://arxiv.org/abs/1908.10084} {Sentence-bert: Sentence embeddings using siamese bert-networks}.
\newblock In \emph{Proceedings of the 2019 Conference on Empirical Methods in Natural Language Processing}. Association for Computational Linguistics.

\bibitem[{Shin et~al.(2020)Shin, Razeghi, Logan~IV, Wallace, and Singh}]{shin-etal-2020-autoprompt}
Taylor Shin, Yasaman Razeghi, Robert~L. Logan~IV, Eric Wallace, and Sameer Singh. 2020.
\newblock \href {https://doi.org/10.18653/v1/2020.emnlp-main.346} {{A}uto{P}rompt: {E}liciting {K}nowledge from {L}anguage {M}odels with {A}utomatically {G}enerated {P}rompts}.
\newblock In \emph{Proceedings of the 2020 Conference on Empirical Methods in Natural Language Processing (EMNLP)}.

\bibitem[{Tan et~al.(2023)Tan, Zhang, and Fu}]{tan2023massive}
Chenmien Tan, Ge~Zhang, and Jie Fu. 2023.
\newblock \href {https://arxiv.org/abs/2311.04661} {Massive editing for large language models via meta learning}.
\newblock \emph{arXiv preprint arXiv:2311.04661}.

\bibitem[{Vrande\v{c}i\'{c} and Kr\"{o}tzsch(2014)}]{wikidata}
Denny Vrande\v{c}i\'{c} and Markus Kr\"{o}tzsch. 2014.
\newblock \href {https://doi.org/10.1145/2629489} {Wikidata: A free collaborative knowledgebase}.
\newblock \emph{Commun. ACM}.

\bibitem[{Wang et~al.(2023{\natexlab{a}})Wang, Haddow, Birch, and Peng}]{wang2023assessing}
Weixuan Wang, Barry Haddow, Alexandra Birch, and Wei Peng. 2023{\natexlab{a}}.
\newblock \href {https://arxiv.org/abs/2310.09820} {Assessing the reliability of large language model knowledge}.
\newblock \emph{arXiv preprint arXiv:2310.09820}.

\bibitem[{Wang et~al.(2023{\natexlab{b}})Wang, Lu, Kong, and Sang}]{wang-etal-2023-towards-alleviating}
Yuhang Wang, Dongyuan Lu, Chao Kong, and Jitao Sang. 2023{\natexlab{b}}.
\newblock \href {https://doi.org/10.18653/v1/2023.findings-acl.270} {Towards alleviating the object bias in prompt tuning-based factual knowledge extraction}.
\newblock In \emph{Findings of the Association for Computational Linguistics: ACL 2023}.

\end{thebibliography}

\clearpage
\appendix

\section{ParaTrex Details}
\label{sec:appendix}
\begin{figure*}[ht!]
\centering
\includegraphics[width=1\linewidth,height=0.55\linewidth]{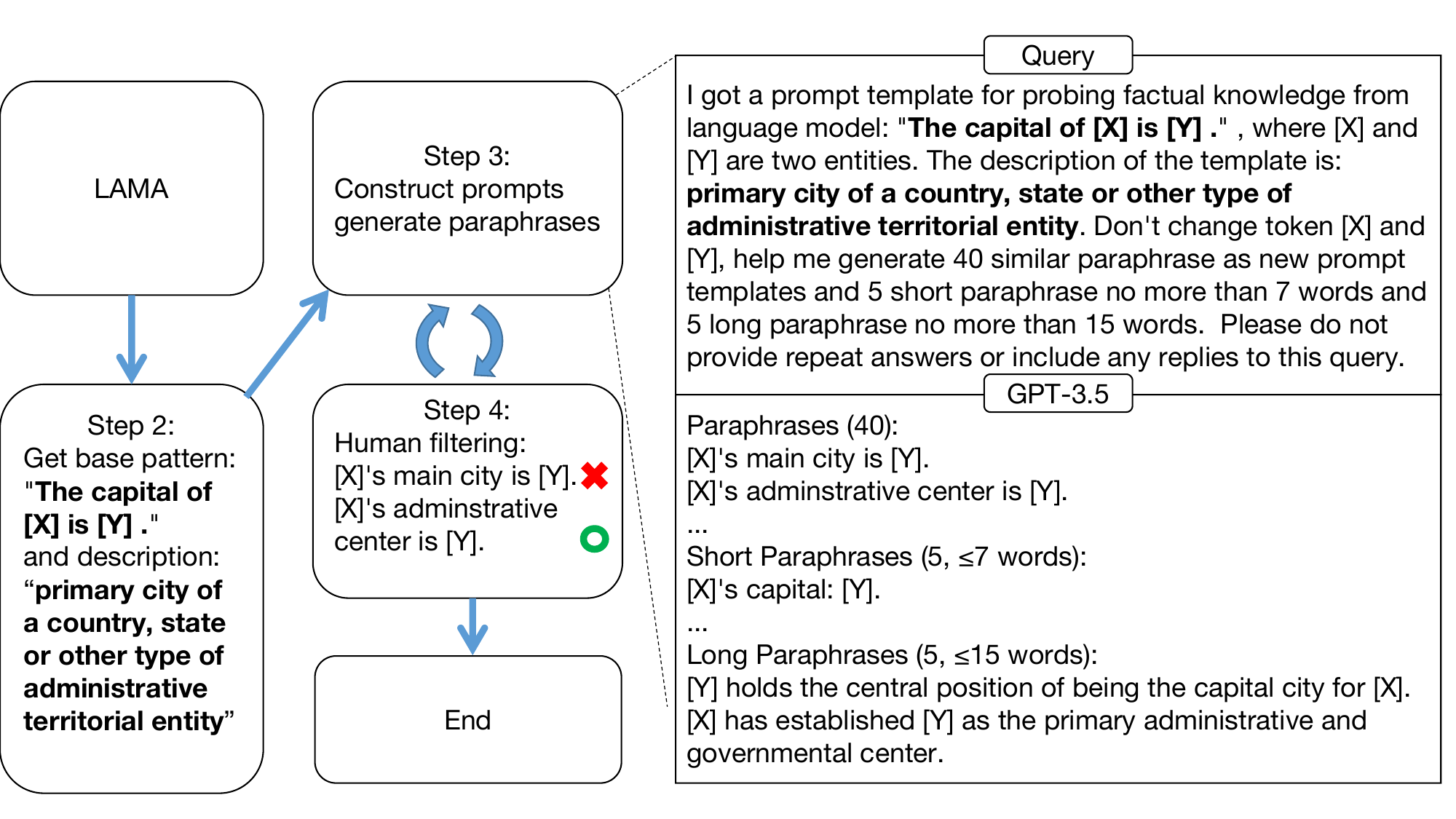}
\caption{Workflow to generate  a paraphrased version of prompt templates in ParaTrex. We exemplify it for the relation `capital of’ in LAMA.}
\label{datacreation}
\end{figure*}
\subsection{ParaTrex: Construction Workflow}
\label{Paratrex:construct}
Fig \ref{datacreation} provides an illustration of the workflow to generate the ParaTrex datasets using large language models.

\subsection{ParaTrex: Exemplary Templates}
\label{Paratrex:details}
Table \ref{paratrex:example} provides an example  of the generated templates in ParaTrex for the relation ``P1376'': ``\textit{Capital\_of}''.

\begin{table*}[ht]
\centering
\small
\begin{tabular}{lcc}
\toprule
Templates                                                        & inhouse split & paraphrase type   \\
\midrule
The capital of {[}Y{]} is {[}X{]} .                              & test          & short paraphrase  \\
{[}X{]} is {[}Y{]}'s capital .                                   & test          & short paraphrase  \\
{[}X{]} serves as {[}Y{]}'s capital .                            & test          & short paraphrase  \\
{[}Y{]}'s capital city is {[}X{]} .                              & test          & short paraphrase  \\
{[}X{]} acts as {[}Y{]}'s capital .                              & test          & short paraphrase  \\
{[}X{]} is the administrative division where the municipality of {[}Y{]} serves as the capital .   & test  & long paraphrase   \\
The governmental seat of {[}Y{]} is located in {[}X{]},  which is the capital city .               & test  & long paraphrase   \\
{[}X{]} holds the status of being the capital city and administrative center of {[}Y{]} .          & test  & long paraphrase   \\
The capital of {[}Y{]} is none other than {[}X{]},  where the government operates .                & test  & long paraphrase   \\
The administrative hub of {[}Y{]} is {[}X{]},  which holds the position of being the capital cit . & test  & long paraphrase   \\
{[}X{]} is the official capital of {[}Y{]} .                     & test          & normal paraphrase \\
The capital city of {[}Y{]} goes by the name of {[}X{]} .        & test          & normal paraphrase \\
{[}X{]} is the designated capital city of {[}Y{]} .              & test          & normal paraphrase \\
{[}X{]} serves as the principal capital city of {[}Y{]} .        & test          & normal paraphrase \\
{[}X{]} is the administrative capital and governmental seat of {[}Y{]} .                           & test  & normal paraphrase \\
{[}X{]} is the principal administrative center of {[}Y{]} .      & test          & normal paraphrase \\
{[}X{]} serves as the capital city and governmental hub of {[}Y{]} .                               & test  & normal paraphrase \\
{[}X{]} holds the official status of being {[}Y{]}'s capital city .                                & test  & normal paraphrase \\
{[}X{]} acts as the administrative capital of {[}Y{]} .          & test          & normal paraphrase \\
{[}X{]} serves as the capital city of {[}Y{]} .                  & test          & normal paraphrase \\
{[}X{]} is the primary governing capital and administrative center of {[}Y{]} .                    & test  & normal paraphrase \\
{[}X{]} is the primary political center of {[}Y{]} .             & test          & normal paraphrase \\
{[}X{]} holds the title of being {[}Y{]}'s capital .             & test          & normal paraphrase \\
{[}X{]} serves as the seat of government for {[}Y{]} .           & test          & normal paraphrase \\
{[}X{]} is the city that serves as {[}Y{]}'s capital .           & test          & normal paraphrase \\
The government of {[}Y{]} is headquartered in {[}X{]},  its capital .                              & test  & normal paraphrase \\
{[}X{]} acts as the political center of {[}Y{]} .                & test          & normal paraphrase \\
{[}X{]} holds the official position of being {[}Y{]}'s capital . & train         & normal paraphrase \\
{[}X{]} serves as the governing center of {[}Y{]} .              & train         & normal paraphrase \\
The capital city of {[}Y{]} is {[}X{]} .                         & train         & normal paraphrase \\
{[}X{]} is the administrative center of {[}Y{]} .                & train         & normal paraphrase \\
The seat of administration in {[}Y{]} is {[}X{]} .               & train         & normal paraphrase \\
The designated capital city of {[}Y{]} is {[}X{]} .              & train         & normal paraphrase \\
The governmental headquarters of {[}Y{]} is located in {[}X{]} . & train         & normal paraphrase \\
{[}X{]} holds the status of being {[}Y{]}'s capital .            & train         & normal paraphrase \\
The government of {[}Y{]} is headquartered in {[}X{]} .          & train         & normal paraphrase \\
{[}X{]} is where the governing body of {[}Y{]} is located .      & train         & normal paraphrase \\
{[}X{]} holds the position of being {[}Y{]}'s capital city .     & train         & normal paraphrase \\
{[}X{]} holds the official governmental seat and capital status of {[}Y{]} .                       & train & normal paraphrase \\
{[}X{]} serves as the governing capital of {[}Y{]} .             & train         & normal paraphrase \\
The capital city of {[}Y{]} is none other than {[}X{]} .         & train         & normal paraphrase \\
The political center of {[}Y{]} is {[}X{]} .                     & train         & normal paraphrase \\
The administrative capital of {[}Y{]} is {[}X{]} .               & train         & normal paraphrase \\
The government headquarters of {[}Y{]} can be found in {[}X{]} . & train         & normal paraphrase \\
{[}X{]} is where the government of {[}Y{]} is based .            & train         & normal paraphrase\\
\bottomrule
\end{tabular}
\caption{Example for the relation ``\textit{Capital\_of}'' in ParaTrex. The original prompt template in LAMA is ``\textit{[X] is the capital of [Y] ."}}
\label{paratrex:example}
\end{table*}

\section{Experiments details and further study}
\subsection{Formal Definitions of Consistency}\label{sec:consistency}
The \emph{consistency} is calculated as the ratio of consistent predictions from different paraphrases with all the paraphrases permutations~\citep{elazar2021measuring,newman2021p1}. Formally, given a set of unordered paraphrase pairs $P_i$ of relation $r_i$, consisting of $n$ distinct prompts, we have a total of $\frac{1}{2}  n(n-1)$ number of permutations. For the $j$-th sample in the $i$-th relation, we define the consistency between all paraphrases as:

\begin{equation} 
    \text{Consistency}_j=\frac{\sum_{p_m, p_n \in P_i}\mathbb{I}[\hat{e}_{ij}^{m}=\hat{e}_{ij}^{n}]}{\frac{1}{2}n(n-1)}
\end{equation}
where $\mathbb{I}$ is the indicator function, $\hat{e}_{ij}^{m}$ and $\hat{e}_{ij}^{n}$ refer to the predicted entity by PLMs from prompt $p_m$ and $p_n$, respectively.

We now give the formal definitions of  \emph{raw-consistency} and \emph{all-consistency}. For the reason of simplicity, we consider the combination of the unique raw prompt template from LAMA, and templates from paraphrased LAMA $p_m \in P_i$, getting $n$ combinations in total. The consistency between raw prompts and paraphrased prompts (\textbf{Raw-Consistency}) will be degraded to:
\begin{equation}
    \text{Raw-Csty}_j=\frac{\sum_{p_m\in P_i, p}\ \mathbb{I}[\hat{e}_{ij}=\hat{e}_{ij}^{m}]}{n}
\end{equation}
Besides, the previous consistency measures only look at the matches between predictions and do not consider strict factual accuracy. However, factual correctness remains a crucial attribute for KBs. Thus, we  additionally measure the consistency over factual correct predictions: 
\begin{equation*}
    \text{Acc-Csty}_j=\frac{\sum_{p_m, p_n \in P_i}\mathbb{I}[\hat{e}_{ij}^{m}=\hat{e}_{ij}^{n}=e_{ij}]}{\frac{1}{2}n(n-1)}
\end{equation*}
, where $e_{ij}$ is the ground truth entity.

\subsection{Training Details}\label{sec:trainingdetails}
We perform all experiments based on BERT-large and RoBERTa-large on the RTX 2080Ti GPUs, which run for about 1 hour to train on one relation. We set the hyperparameter $\lambda_{me}, \lambda_{kld}$ to be 0.2. $w_{true}$ and $w_{false}$ are set to be simply -1 and 1. For adapters, we take the hidden state to be 256 dimensions. All other hyperparameters (including the random seed) are set as default in \cite{LIU2023}.

\subsection{Significance Test Details}\label{sec:sg_test}
We perform the Paired sample T-test and the Wilcoxon Signed-Ranked test on the results from all 25 relations between adapters and our UniArk to test the significance after performing UniArk. We also apply different seeds (20, 30, 50) and perform a t-test among the average results to test whether the results are significant for different runs. The results of the p-values are shown in Table \ref{sigtest}, where cst refers to the consistency, pt, pr, and lm refer to the ParaTrex, ParaRel, and LAMA datasets respectively.

Overall, we can observe that the p-values of all consistency and out-of-domain f1 scores are smaller than 2.5e-2, strongly suggesting that UniArk makes significant improvements over the baseline adapters both with the normally distributed assumption or not. On the contrary, all results in the in-domain f1 scores are bigger than 5e-2, indicating the non-significance of the decrease/increase in in-domain quality. This proves that UniArk makes significant improvements over the out-of-domain generation and both biases while maintaining its performance in the in-domain settings.

\subsection{Details after removing the semantic overlapped relations}\label{appendix:semantic overlap}
\begin{table}[ht]
\centering\small
\setlength{\tabcolsep}{0.6ex}
\begin{tabular}{clcccccc}
\toprule
\multirow{2}{*}{Model} &
\multirow{2}{*}{Method} &
  \multicolumn{3}{c}{ParaTrex} &
  \multicolumn{3}{c}{ParaRel} \\
\cmidrule(lr){3-5}\cmidrule(lr){6-8} 
& &  \multicolumn{1}{r}{raw} &
  \multicolumn{1}{c}{all} &
  \multicolumn{1}{c}{acc} &
  \multicolumn{1}{r}{raw} &
  \multicolumn{1}{c}{all} &
  \multicolumn{1}{c}{acc} \\
\midrule
\multicolumn{1}{c}{\multirow{2}{*}{\makecell{BERT\\-large}}} 
 &
  Adapters &
  68.0 &
  61.9 &
  \multicolumn{1}{c}{47.6} &
  73.3 &
  66.2 &
  50.6 \\
 &
  \textbf{+ UniArk} &
  \textbf{{\ul74.2}} &
  \textbf{{\ul69.7}} &
  \multicolumn{1}{c}{\textbf{{\ul 52.5}}} &
  \textbf{{\ul77.3}} &
  \textbf{{\ul72.1}} &
  \textbf{{\ul54.4}} \\
\bottomrule
\end{tabular}
\setlength{\belowcaptionskip}{-0.2cm}
\caption{Main results for consistency on ParaTrex and ParaRel after \textbf{removing the semantic overlapped relations}. Significantly improved results are underlined.}
\label{append:consistency}
\end{table}
The problem of semantic overlapping when measuring the consistency for the factual probing task was pointed out by \cite{hagstrom-etal-2023-effect}. This problem refers to the scenarios when the model is allowed to choose between semantically close answer alternatives, but only one of these is accepted as a correct answer. For instance, relation P101 \textit{field-of-work} contains both biology and science, and relation P19 \textit{born-in} contains both Glasgow and Scotland, where either predictions is supposed to be correct. We follow the results from \cite{hagstrom-etal-2023-effect}, remove those 12 relations under the risk of semantic overlapping, and report the experiment results between UniArk and baseline Adapters again in Table \ref{append:consistency}. Here we observe a similar trend of improvements with the full results shown in Table \ref{Table4}, suggesting that semantic overlap minimally influences the enhancements attributed to UniArk.

\subsection{Scaling Study}
\label{scalestudy}

\begin{figure*}[t]
\centering\small
\includegraphics[width=0.7\linewidth,height=0.35\linewidth]{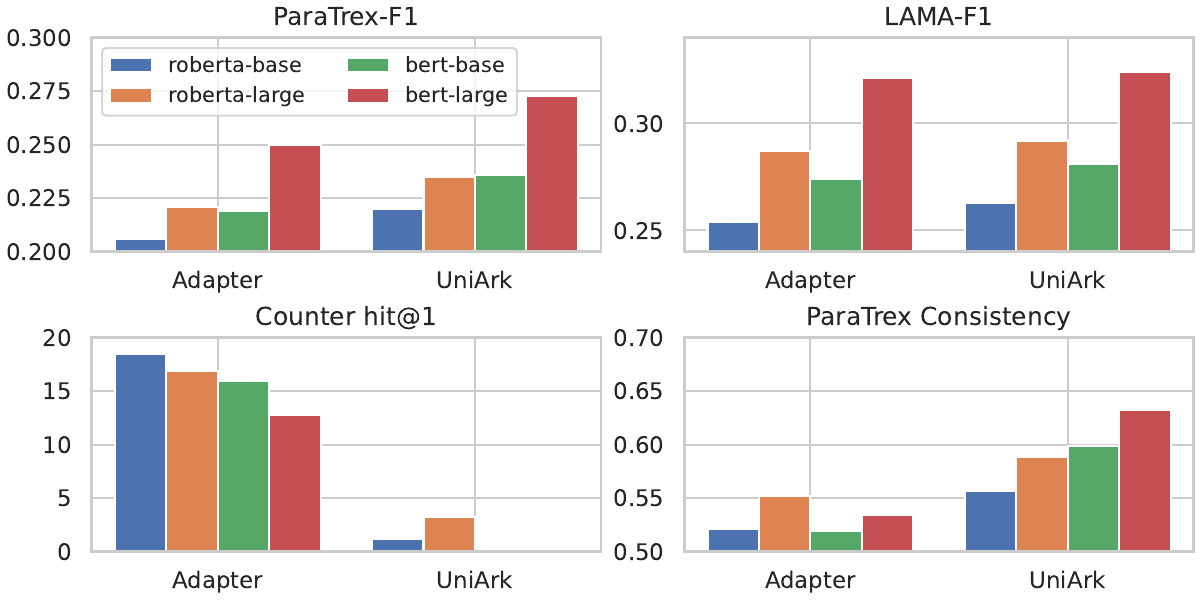}
\caption{Sscaling results between adapters and UniArk with different scales of models.}
\label{scaling}
\end{figure*}

We want to answer the question of whether the results of UniArk are scalable for models with more parameters. Fig \ref{scaling} presents a comparison of F1 scores, counterfactual accuracy and  consistency between BERT-base, BERT-large, RoBERTa-base, and RoBERTa large.  The results demonstrate that UniArk performs consistently better for both extraction performance and inherent bias. We also observe consistently better results for larger models among all settings. We therefore conclude that (1) The performance for extracting knowledge and bias can be scaled by the size of LMs. (2) The bias mitigation and performance boost from the UniArk framework can also be observed among all sizes of models (3) For bias mitigation, small models are able to be more unbiased and robust through the UniArk framework.

\subsection{Details for Qualitative Study}\label{sec:quality study}

\begin{table*}[ht]
\small
\centering
\begin{tabular}{llllllll}
\toprule
Paired T-test &
  \textbf{ood\_f1\_pt} &
  \textbf{ood\_f1\_pr} &
  \textbf{all\_cst\_pt} &
  \textbf{all\_cst\_pr} &
  \textbf{acc\_cst\_pt} &
  \textbf{acc\_cst\_pr} &
  \textbf{id\_lm\_f1} \\
\midrule
BERT               & 1.36e-04 & 3.19e-03 & 1.26e-06 & 7.82e-06 & 2.40e-05 & 6.20e-05 & 6.26e-01 \\
RoBERTa            & 7.35e-04 & 9.39e-03 & 2.19e-03 & 1.69e-04 & 7.28e-03 & 2.92e-03 & 4.61e-01 \\
\midrule
Wil rank Test &          &          &          &          &          &          &          \\
\midrule
BERT               & 1.83e-05 & 3.78e-03 & 1.19e-07 & 4.17e-07 & 2.56e-06 & 8.34e-07 & 5.37e-02 \\
RoBERTa            & 7.50e-05 & 1.15e-02 & 2.17e-04 & 1.51e-05 & 2.87e-04 & 3.29e-04 & 5.65e-02 \\ 
\midrule
T-Test &          &          &          &          &          &          &          \\
\midrule
BERT               & 1.06e-04 & 4.80e-03 & 6.13e-04 & 5.02e-04 & 6.09e-05 & 2.73e-04 & 5.03e-02 \\
RoBERTa            & 1.48e-03 & 1.23e-02 & 5.21e-03 & 3.63e-03 & 1.16e-03 & 1.09e-03 & 6.65e-02 \\ 
\bottomrule
\end{tabular}
\caption{Significance test between adapter baseline and UniArk over 41 relations for f1 score and 25 relations for consistency (cst) on ParaTrex (pt) and ParaRel (pr).}
\label{sigtest}
\end{table*}

\begin{table*}[t]
\centering\small
\setlength{\tabcolsep}{0.7ex}
\begin{tabular}{llcc}
\toprule
\multicolumn{2}{c}{Inputs (Subject: Vesanto, Object: Finnish)}                                                  & \multicolumn{2}{c}{Predictions}                        \\ \midrule
\multicolumn{1}{l|}{Type} & Prompt template                          & \multicolumn{1}{c}{Adapter-Tuning} & UniArk        \\ \midrule
\multicolumn{1}{l|}{raw}  & The official language of {[}X{]} is {[}MASK{]}.                                      & \textbf{Finnish}                    & \textbf{Finnish} \\ \midrule
\multicolumn{1}{l|}{\multirow{6}{*}{paraphrased}} & {[}X{]} designates {[}MASK{]} as the official language . & Italian & \textbf{Finnish} \\
\multicolumn{1}{l|}{}     & {[}X{]} has {[}MASK{]} as its official language .                                    & It                                  & \textbf{Finnish} \\
\multicolumn{1}{l|}{}     & {[}MASK{]} has been declared as the recognized language in {[}X{]} .                 & Finland                             & \textbf{Finnish} \\
\multicolumn{1}{l|}{}     & In {[}X{]}, {[}MASK{]} is acknowledged as the prescribed language by the government. & It                                  & Finland          \\
\multicolumn{1}{l|}{}     & The officially recognized language in {[}X{]} is {[}MASK{]} .                        & Italian                             & Italian          \\
\multicolumn{1}{l|}{}     & {[}X{]} recognizes {[}MASK{]} as its official language .                             & Italian                             & \textbf{Finnish} \\
\bottomrule
\end{tabular}
\setlength{\belowcaptionskip}{-0.2cm}
\caption{LM prediction examples from the raw inputs in LAMA and the diverse paraphrased prompts in ParaTrex.}
\label{case-pv}
\end{table*}

\begin{table*}[ht]
\centering\small
\setlength{\tabcolsep}{0.7ex}
\begin{tabular}{ccccc}
\toprule
\multirow{2}{*}{Method}                 & \multirow{2}{*}{Input}                      & \multicolumn{3}{c}{Subject=``Sorengo''} \\ 
\cmidrule(lr){3-5}
& & Top 1 & Top 2 & Top 3 \\
\midrule
\multirow{4}{*}{UniArk} & \multirow{2}{*}{raw}                                                & \textbf{Italian} & Finnish     & Swedish     \\
                       &                             & 0.1213     & 0.1152             & 0.1125         \\
                           & \multirow{2}{*}{\begin{tabular}[c]{@{}c@{}}subject \\ masked\end{tabular}} & Polish        & German & Greek \\
                       &                             & 0.0423     & 0.0421             & 0.0421     \\ \midrule
\multirow{4}{*}{MeCoD} & \multirow{2}{*}{raw} & Finnish      & Swedish               & Norwegian         \\
                       &                             & 0.1322     & 0.1232             & 0.1041      \\
                           & \multirow{2}{*}{\begin{tabular}[c]{@{}c@{}}subject \\ masked\end{tabular}} & French        & Danish   & Armenian \\
                       &                             & 0.1153     & 0.1051             & 0.0995    \\ \midrule
\multirow{4}{*}{LAMA}  & \multirow{2}{*}{raw} & Portuguese     & English     & Spanish      \\
                       &                             & 0.116     & 0.1146             & 0.1125    \\
                           & \multirow{2}{*}{\begin{tabular}[c]{@{}c@{}}subject\\ masked\end{tabular}}  & English & French    & Spanish    \\
                       &                             & 0.1111     & 0.1079             & 0.1016     \\ \bottomrule
\end{tabular}
\setlength{\belowcaptionskip}{-0.2cm}
\caption{Case study on top-3 objects and their logits extracted by LMs through the original prompt template.}
\label{case-pp}
\end{table*}

We perform two specific case studies to better understand how mitigating the studied biases helps to improve the knowledge extraction results. Firstly, in Table~\ref{case-pp} we present cases showcasing  how the models make the incorrect prediction due to the biased object likelihood. PLMs are asked for the official language of a specific item using the prompt:``\textit{The official language of [sub] is [obj].}". The last row shows the results for the vanilla LMs without being tuned and thus suffering from high object likelihood such as \textit{English} and \textit{Spanish}. The logits of objects \textit{English} and \textit{Spanish} of LAMA methods are close, showing that the model is not confident with its predictions and may guess from the object likelihood from templates. The SoTA model MeCoD still gives the wrong answer since they apply an unreliable neural gate to automatically classify which object to be debiased. For instance, MeCoD successfully smooths the high counterfactual logit for the  word \textit{English} but causes the model to underfit this object so that it cannot recall the correct object Italian and thus make an incorrect prediction with a high logit. In contrast, UniArk is capable of making accurate predictions with higher logits while having an unbiased prediction distribution under subject-masked inputs, showing that UniArk provides more confident answers without the impact of the prior distribution from  prompt templates.

Table \ref{case-pv} presents an example of the consistency study. We provide an instance where adapter-tuning and UniArk are both correct on the original prompts. We randomly sample several paraphrased cases from ParaTrex. The results suggest that the baseline fails to produce correct answers when meeting syntactically and lexically diverse prompt templates. The second and fourth rows of paraphrased prompt templates are examples 
for the different syntic variants while the first and the last rows of paraphrased templates show more lexically complicated prompts. Our UniArk model gives mostly  consistent outputs in those cases, although it may make some mistakes. Additionally, we can observe from the results that UniArk  maintains a robust behaviour in outputting language objects instead of stopwords like ``\textit{it}''. This shows that the UniArk models are more robust on various prompt templates after debiasing.

\subsection{Details for the Error Analysis}
\label{error_ans}
To have a comprehensive understanding of what kinds of errors UniArk made, we random sample 50 wrong predictions among 4283 error samples in relation P37 ``\textit{Official\_Languages}''. Results are shown in Table \ref{error_anlysis}.  
\begin{table*}[ht]
\centering\small
\setlength{\tabcolsep}{0.7ex}
\begin{tabular}{llllcc}
\toprule
\multirow{2}{*}{Error Type} & \multirow{2}{*}{N} & \multicolumn{4}{c}{Example} \\
\cmidrule(lr){3-6}
& & Subject & Prompt & Golden & Prediction \\
\midrule
Unknown Case 
& 23
& Azad Kashmir
& Azad Kashmir bestows official language status upon [Y] .
& Urdu
& English
\\
Spelling Error &
2 &
Melitopol &
[Y] holds the official language designation of Melitopol .&
Ukrainian&
Ukraine
\\
Pronouns
& 4
& Malax
& [Y] is officially recognized as the language of [X] .
& Finnish
& It
\\
Multiple Correct Answers &
21&
ASEAN &
The designated official language of ASEAN is [Y] . &
Thai &
Indonesia
\\
\bottomrule
\end{tabular}
\caption{Types of errors appeared in UniArk on LAMA and ParaTrex test datasets}
\label{error_anlysis}
\end{table*}

\begin{figure*}[t]
\centering
\includegraphics[width=1\linewidth]{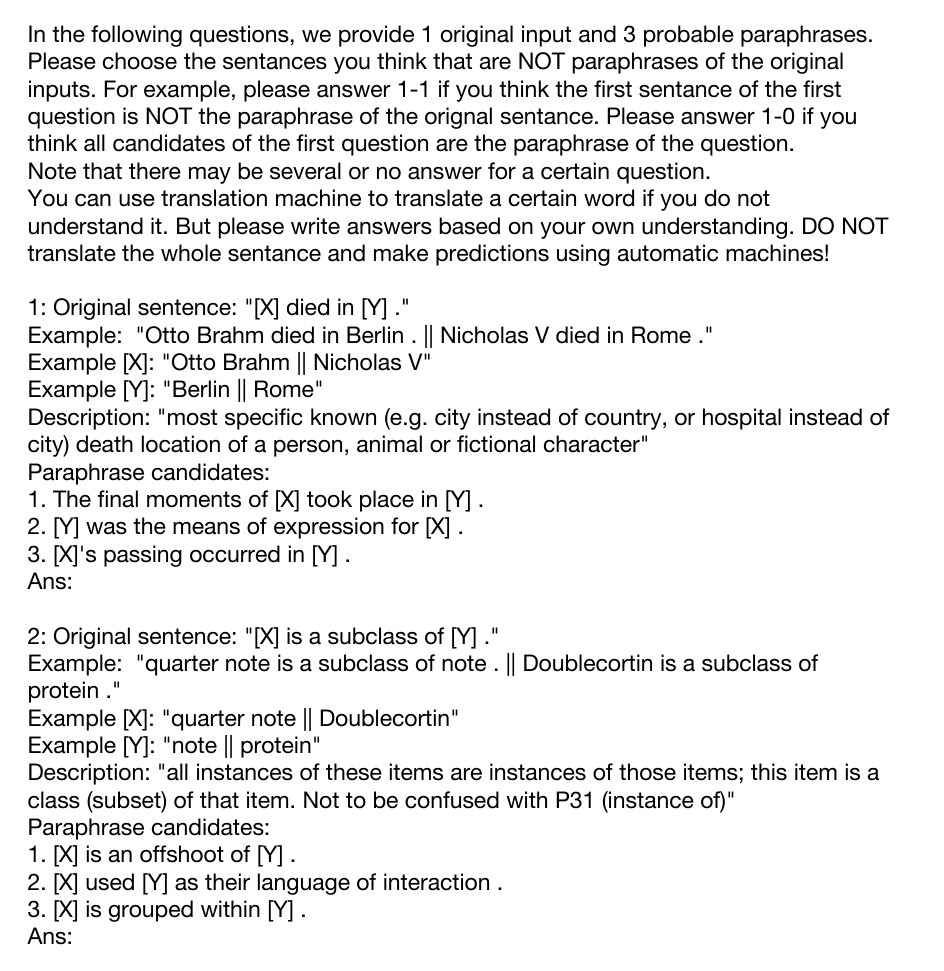}
\caption{Example of the questions for human evaluation}
\label{paratrex:he}
\end{figure*}

\end{document}